\documentclass[submission,copyright,creativecommons]{eptcs}

\providecommand{\event}{FMAS 2024} 

\usepackage{iftex}

\ifpdf
  \usepackage{underscore}         
  \usepackage[T1]{fontenc}        
\else
  \usepackage{breakurl}           
\fi

\usepackage[dvipsnames]{xcolor}
\usepackage[english]{babel}
\usepackage{lettrine}
\usepackage{enumitem}
\usepackage{amsfonts}
\usepackage{listings}
\definecolor{codegreen}{rgb}{0,0.6,0}
\usepackage{url}
\usepackage{graphicx}
\usepackage{dirtytalk}
\usepackage{float}
\usepackage{parskip}
\usepackage{caption}
\usepackage{subcaption}
\usepackage{amsmath}
\usepackage{mathtools}
\usepackage{setspace}
\usepackage{hyperref}
\usepackage{makecell}
\usepackage[skins]{tcolorbox}
\usepackage[intoc]{nomencl}
\usepackage[font=small,labelfont=bf]{caption}
\usepackage{pdfpages}
\usepackage{multirow}
\usepackage{amssymb}
\usepackage{cprotect}
\usepackage{tikz}

\definecolor{codegreen}{rgb}{0,0.6,0}
\usepackage[T1]{fontenc}
\usepackage[scaled=0.85]{beramono}
\makeatletter
\newcommand\monottfamily{%
  \def\fvm@Scale{0.75}
  \fontfamily{fvm}\selectfont
}
\makeatother

\usepackage{listings}
\usepackage{colortbl}
\usepackage{color}
\lstdefinelanguage{csp}{
  keywords=[1]{assertion, untimed, csp, refines, timed, is, deadlock, free, reachable, associated},
  keywords=[2]{nametype,channel,Timed,if,then,else, timed_priority,WAIT,SKIP,STOP,USTOP,!,?,[],->,[],::,[|,|],\{|,|\}},
  sensitive=true, 
  breaklines=true, 
  comment=[l]{//},
  morecomment=[s]{/*}{*/},
  morecomment=[l]{--},
  commentstyle=\color{Green},
  moredelim=[is][\color{Plum}\bfseries]{/@}{@/},
}

\usepackage{float}
\newfloat{lstfloat}{htbp}{lop}
\floatname{lstfloat}{Listing}

\usepackage{comment}

\lstdefinelanguage{csp}{
  keywords=[1]{assertion, untimed, csp, refines, timed, is, deadlock, free, reachable, associated},
  keywords=[2]{nametype,channel,Timed,if,then,else, timed_priority,WAIT,SKIP,STOP,USTOP,!,?,[],->,[],::,[|,|],\{|,|\}},
  sensitive=true, 
  breaklines=true, 
  comment=[l]{//},
  morecomment=[s]{/*}{*/},
  morecomment=[l]{--},
  commentstyle=\color{codegreen},
  moredelim=[is][\color{Plum}\bfseries]{/@}{@/},
}

\newcommand{\RC}[1]{{\sf #1}}

\title{Model Checking and Verification of \\
Synchronisation Properties of Cobot Welding}
\author{Yvonne Murray
\institute{Pioneer Robotics AS\\ Dept.\ of Mechatronics, University of Agder \\ Norway}
\email{ym@pioneer-robotics.no}
\and
Henrik Nordlie
\institute{Robotics Group, Faculty of Science \& Technology\\
Norwegian University of Life Sciences (NMBU)\\ Norway}
\and
David A. Anisi
\institute{Dept.\ of Mechatronics, University of Agder \\ Robotics Group, Faculty of Science \& Technology\\
Norwegian University of Life Sciences (NMBU)\\ Norway}
\and
Pedro Ribeiro 
\institute{Dept.\ of Computer Science\\University of York\\ UK}
\and 
Ana Cavalcanti
\institute{Dept.\ of Computer Science\\University of York\\ UK}
}

\def\titlerunning{Model Checking and Verification of 
Synchronisation Properties of Cobot Welding}
\def\authorrunning{Y. Murray, H. Nordlie, D.A. Anisi, P. Ribeiro \& A. Cavalcanti}

\begin{document}

\maketitle


\begin{abstract}
This paper describes use of model checking to verify synchronisation properties of an industrial welding system consisting of a cobot arm and an external turntable. The robots must move synchronously, but sometimes get out of synchronisation, giving rise to unsatisfactory weld qualities in problem areas, such as around corners. These mistakes are costly, since time is lost both in the robotic welding and in manual repairs needed to improve the weld. Verification of the synchronisation properties has shown that they are fulfilled as long as assumptions of correctness made about parts outside the scope of the model hold, indicating limitations in the hardware. These results have indicated the source of the problem, and motivated a re-calibration of the real-life system. This has drastically improved the welding results, and is a demonstration of how formal methods can be useful in an industrial setting.  
\end{abstract}

\section{Introduction}
\label{sec:intro}

\begin{figure}[t]
    \centering
    \includegraphics[scale=0.6]{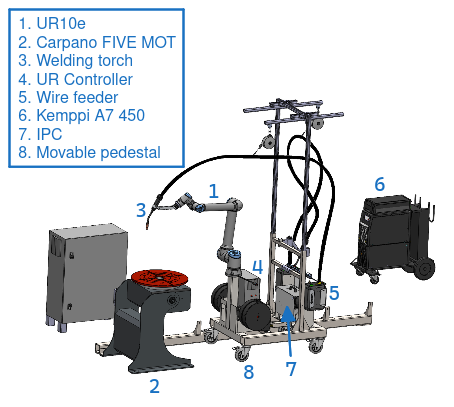}
    \caption{The IntelliWelder system with the different components marked by number~\cite{NordlieThesis}.}
    \label{fig:IntelliWelder}
\end{figure}

Robotic welding is commonly used in industrial workshops to increase efficiency and repeatability, and reduce dangerous and ergonomically straining work for human welders~\cite{KahP2015Raws}. To address the needs of small and medium-sized enterprises (SMEs), which produce a large variety of products in small quantities, the welding system must be easy and fast to re-program, and highly flexible. To this end, Pioneer Robotics have developed the IntelliWelder M06~\cite{IntelliWelder_Online}, a flexible and light-weight welding system consisting of a Universal Robots (UR)~UR10e cobot~\cite{colgate1996cobots} equipped with a welding torch  and a Carpano FIVE MOT turntable serving as an external axis (EXAX). Fig.~\ref{fig:IntelliWelder} shows the components of the IntelliWelder.

\begin{figure}[t]
    \centering
    \includegraphics[scale=0.1]{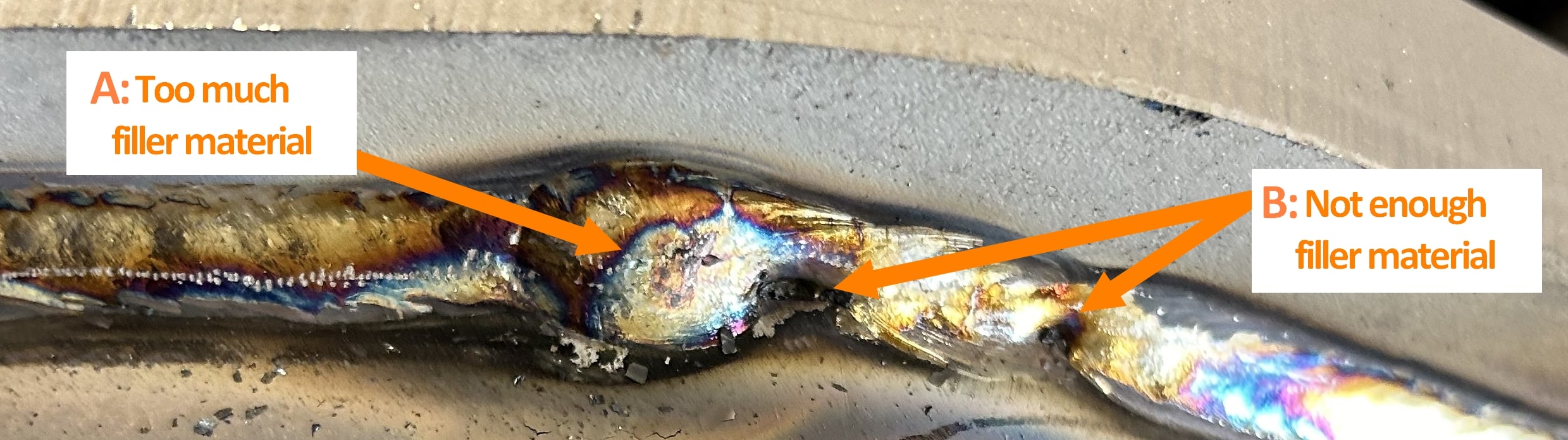}
    \caption{Typical welding issue where there is buildup of filler material (A) and coverage is not sufficient (B), creating an irregular and weakened weld.}
    \label{fig:WeldProblem}
\end{figure}

The main challenge in the operation of the IntelliWelder M06 has been to get a high quality, continuous weld in difficult areas, such as around corners. To get the best results, it is important that the welding robot and the turntable move continuously in a \emph{synchronous} fashion. By using synchronous welding, it is possible to achieve a continuous weld of high quality, ensuring the weld is not jagged and interrupted. When the synchronisation does not work properly, the welding gun does not move forward in an even motion, and can move too fast or too slow. Moving too fast does not give the metal and filler enough time to heat up and weld together, while moving too slow results in build-up of filler material. Both of these problems can be seen in the weld depicted in Fig.~\ref{fig:WeldProblem}.

This experience report describes how we have addressed some challenges faced when the UR robot and the EXAX move synchronously while welding. Relevant parts of the system have been modelled in RoboChart~\cite{miyazawa_robochart_2019}, a domain-specific language for modelling and verification of robotic systems. Using our RoboChart model, key synchronisation properties have been verified using the refinement model checker FDR~\cite{FDR_online}. Details omitted here are in~\cite{NordlieThesis}; the work demonstrates how formal verification can be used in industry, and how the results can be used to localise the error source, leading to system improvements.

Previous research on multi-robot welding include~\cite{StarkeGünther2016Sasi}, which focuses on nominal trajectory planning and self-coordination, and~\cite{XiongJiahao2021TOaS}, which studies trajectory smoothing in a dual-robot collaborative welding system. Neither of these lines of work use formal methods or model checking. Closer to our research, the work in~\cite{formal_welding} combines graphic and formal methods to analyse collaborative behaviour such as deadlock and equivalence properties. None of these works, however, consider the issue of correct time synchronisation during multi-robot execution like we do in our case study. 

The rest of this paper is organised as follows.  In Section~\ref{sec:FV_MC}, we motivate our use of formal methods and model checking. In Section~\ref{sec:intelliwelder}, we detail the system architecture and requirements, before the model is presented in Section~\ref{sec:Modelling}. In Section~\ref{sec:modelchecking}, we present the verification results and their practical implications. Finally, in Section~\ref{sec:conclusion}, we conclude, describe ongoing work, and suggest further work.



\section{Formal Verification and Model Checking}
\label{sec:FV_MC}

To find and mitigate faults and undesired behaviour in robotic systems, they are traditionally subject to testing, including simulation before deployment. For real-world, complex robotic systems, however, it is impossible to test every possible scenario and input sequence. Moreover, even if a fault is discovered, error source localisation remains a challenge. In this setting, formal verification methods are a useful supplement. Model checking~\cite{book_model_checking,goos_model_1997} is a formal method to verify that given properties are fulfilled, regardless of inputs. If a property does not hold, model checking  provides a counterexample that can pinpoint the cause of error. Adopting such methods is valuable in the design of real, industrial systems.

RoboTool~\cite{miyazawa_robochart_2019} is a suite of plugins for the Eclipse IDE supporting use of the RoboStar framework~\cite{cavalcanti_robostar_2021}. 
Our previous work on verification of an industrial control system~\cite{murray_safety_2022} using RoboStar has shown its proficiency and strengths. In RoboStar, a key artefact is a  RoboChart~\cite{miyazawa2016robochart} model that reflects the real system design. Once this  model is created, assertions for the selected properties can be written and verified using the CSP process algebra and its model checker, FDR~\cite{gibson-robinson_fdr3_2014}.

As our use case considers an already existing IntelliWelder system, we need to alter the idealised workflow of RoboStar~\cite{cavalcanti_robostar_2021} by effectively "reverse engineering" the RoboChart model from the existing system. 

\section{IntelliWelder and Synchronous Welding}
\label{sec:intelliwelder}

In this section, we describe our case study:~its architecture~(Section~\ref{section:architecture}), software~(Section~\ref{section:code}), and requirements~(Section~\ref{sec:specs}), identifying the problem we are addressing with model checking. 

\subsection{System Architecture}
\label{section:architecture}

\begin{figure}[h]
    \centering
    \includegraphics[width=1\textwidth]{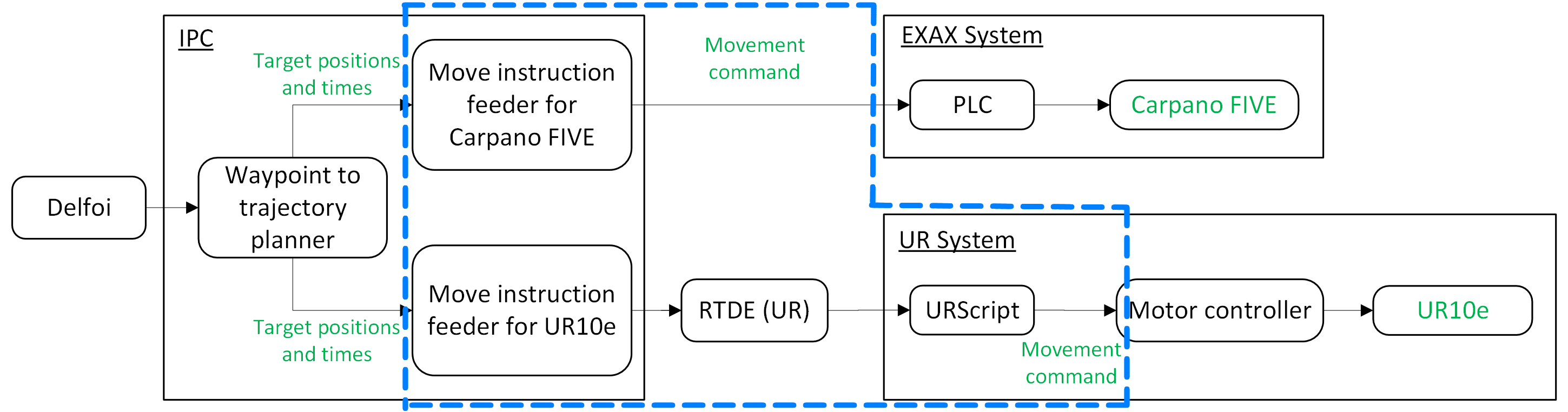}
    \caption{System architecture of the IntelliWelder. The blue dotted line indicates the scope of the RoboChart model:~part of the Industrial PC (IPC), the Real-Time Data Exchange (RTDE) for the UR robot, and the URScript.}
    \label{fig:SystemArchitecture}
\end{figure}

An illustration of the IntelliWelder's architecture can be seen in Fig.~\ref{fig:SystemArchitecture}.
To realise synchronous welding, Delfoi offline robot programming software from Visual Components~\cite{Delfoi_online} is used for creating waypoints and welds in a 3D layout consisting of the UR10e~\cite{UR10e-online}, the Carpano FIVE turntable, and the workpiece to be welded. Welds are created simply by selecting an edge on the workpiece CAD model. Delfoi then creates waypoints for both the UR robot and the Carpano turntable, so that each waypoint for the robot has a corresponding waypoint for the turntable, creating nominally synchronised movements.

The waypoint paths generated in Delfoi are transferred to the Industrial PC~(IPC). The waypoints are then processed to convert them into trajectories based on the desired forward welding speed and other welding parameters. 
The waypoints for the external axis remain unchanged and are converted into a trajectory, while the waypoints for the UR robot are sampled at a higher resolution before being turned into its trajectory. Consequently, the UR robot has more waypoints to process than the Carpano FIVE. Once the trajectories are generated, the IPC sends them as movement requests for the system to execute.

The Carpano FIVE is controlled by a Programmable Logic Controller~(PLC) that receives movement commands from the IPC. These commands can be based on position and velocity, or just velocity. The PLC then regulates movement using a PID. The Real-Time Data Exchange~(RTDE) synchronises external applications with the UR robot~\cite{rtde-ur}. It relays messages from the IPC to the UR robot via a TCP/IP connection. 
The UR robot controller executes URScript applications and manages movement via a PID. 

\subsection{Existing  URScript code}
\label{section:code}

In the current implementation, every time a movement request for the UR robot arrives, the URScript code runs on the UR controller. Required variables are read from the registers, updated by the RTDE, and the code checks if the target time for the next waypoint has already passed (that is, the robot is behind schedule). If so, the code logs a warning and continues to the next target.

Next, the URScript decides which movement type is preferable to reach the next waypoint, based on variables like blend radius, offset, joint velocities, and whether the next movement involves a sharp turn. The script selects between \RC{MoveJ}, \RC{MoveL} and \RC{MoveP}, which are standard UR robot movements described in the user manual~\cite{URmanual}. However, if none of the standard movement types are suitable, a custom-made function called \RC{MoveL\_with\_t} is used for the movement. The custom URScript function \RC{MoveL\_with\_t} uses the standard \RC{MoveL}
command with the next target pose and next target time as arguments. In that way, the UR robot can calculate the necessary velocity to reach the next target within the target time, using maximum acceleration. The reason for this being a fallback solution, only used when the standard movements are not feasible, is that it does not include a blend radius. 

With a blend radius, we ensure that when the UR robot is within a given distance of the waypoint, it starts moving towards the next waypoint instead of completely finishing the move to the current waypoint. So instead of coming to a brief stop at the waypoint, it keeps moving towards the next, giving a smoother transition. Lack of a blend radius results in a jagged movement that is not ideal for welding.

The next section describes the requirements that the design just presented is expected to satisfy. 

\subsection{Requirements}
\label{sec:specs}

We present here both system requirements~(in Section~\ref{section:system-reqs}), and requirements specifically for the components that we model as described in Figure~\ref{fig:SystemArchitecture}~(in Section~\ref{section:model-reqs}).

\subsubsection{System-Wide Requirements}
\label{section:system-reqs}

There are several requirements for the IntelliWelder system as a whole, discussed in detail in~\cite{NordlieThesis}. The most important of these requirements are the following two:
\begin{enumerate}
    \item The welding torch must always stay in an area defined by a maximum deviation from the weld frame. This includes both position and orientation.
    \item The welding torch must always move forward in the weld frame with a speed that is within a given maximum deviation of the desired forward weld speed.
\end{enumerate}
These requirements 
need to be refined 
into specific requirements for Delfoi, the IPC planner, the calculation of arguments for the robot commands, and the execution of movements. Additionally, they 
expand to include requirements related to information communication and code execution time.

\subsubsection{Model Requirements}
\label{section:model-reqs}

With the system wide requirements in mind, the following requirements for the modelled component~(see Figure~\ref{fig:SystemArchitecture}) can be obtained, as described in detail in~\cite{NordlieThesis}:

\noindent%
{\bf R1} The component should detect events that imply that the system is out of sync.
    
\noindent
{\bf R2} For each movement request received from the IPC, the corresponding robot should receive a movement command unless the system is out of sync.


The requirements R1 and R2 above are the properties we verify using model checking. If some of the assertions fail, it can help to pinpoint existing mistakes in the software. If all assertions pass, it indicates that some of the assumptions made on the component's context and the hardware are invalid.

To check the hardware, two different cases are evaluated:~one where it is impossible for the robot to receive a waypoint that is already in the past~(nominal case), and one where that is possible~(realistic case). If the model checking results vary between the two, for example, if the assertions pass in the nominal case~(which assumes the hardware is able to keep to the planned trajectory) but fail in the more realistic case, it is an indication that assumptions about velocities, accelerations, and perfect move execution, made on the real-life system, are inaccurate. We recall that the system does present a problem. So, if the problem is not present when the hardware executes the planned trajectory, then we can conclude that our assumptions about the hardware are not satisfied, and so, inaccurate. 

In the next section, we present the RoboChart model we use to carry out our verification. 
\section{Modelling in RoboChart}
\label{sec:Modelling}

Our RoboChart model reflects the system architecture already described, and the existing code and specifications.  
Any possible communication delays are assumed to be handled separately and are hence negligible for our purposes here. The components modelled receive movement requests as inputs. 
Thus, trajectory planning is outside of the model's scope and the feasibility of planned trajectories is assumed. 

As previously noted, the number of waypoints differs for the UR robot and the Carpano FIVE, so both are expected to receive and execute commands concurrently and independently. In terms of control flow, the model's scope extends until the point where these movement commands are initiated for the Carpano FIVE and the UR robot. From that point, they handle the execution of movements. We expect and assume that the actual execution of movement commands by the UR robot and Carpano FIVE is correct and, therefore, that is also beyond the model's scope.

The definition of the model's scope reflects the fact that our goal is to check that our use of the Carpano FIVE and UR robot commands is appropriate.  Therefore, in the RoboChart model, these commands are captured as services of the robotic platform, which we do not further specify.  

The component modelled is responsible for selecting the most appropriate movement type for each request. It also detects if the system is out of sync, meaning the target time for a movement request has already passed, resulting in a negative time budget. 
In the model, this is indicated by the occurrence of an \RC{out-of-sync} event, and is considered a critical failure.

Next, we justify the abstractions and simplifications made in the RoboChart model~(Section~\ref{section:abstractions-simplifications}), and then present the RoboChart model itself~(Section~\ref{section:robochart}).

\subsection{Abstractions and Simplifications}
\label{section:abstractions-simplifications}

It is well-known that model checking eventually encounters state-explosion problems. 
To keep the complexity and verification time at bay, the following abstractions and simplifications have been made.

\paragraph{Reduction of number of joints:}
The Carpano FIVE turntable has two joints, one for tilting and one for turning. The IntelliWelder, however, only uses the turning axis during the synchronous welding. Thus, this is the only axis that is considered in the model. The UR robot has six rotational joints, but it is modelled with only two. Although this selection can be made arbitrarily, selecting one from the first three joints (to represent position) and the other from the last three joints (to represent orientation) is advocated. By modelling two axes, the model still captures potential problems related to multiple joint values.  Extending the model to include six axes affects the computational complexity of the model, as it would lead to additional parameters of operations (explained in the next section). These operations, however, are not further specified as they represent services that are out of the scope of our verification. So, additional parameters are not relevant for the verification, but just the fact that they are available. 

\paragraph{Limited value ranges:}
To decrease the computation time and complexity, the value ranges of the variables of type \RC{real}, recording distance, are limited. To cover both negative, positive, and zero-values, the integer range \RC{[-1..1]} has been chosen. 
Similarly, the \RC{int} variables recording discrete time are limited to the two ranges \RC{[0..2]} and \RC{[-1..1]}. With the first range, with only positive values, the assumption is that the UR robot and EXAX are never so late that the next waypoint is already in the past. When a negative value -1 is included, we can check whether the goal time for the next waypoint has passed.
Lastly, the variables recording the current waypoint for the UR robot and the turntable, of datatype \RC{nat}, are limited to the ranges \RC{[0..3]} and \RC{[0..1]}, respectively. So, the maximum number of waypoints for the UR robot are 4 and for the turntable 2, capturing that the number of waypoints can vary in the real system.

\paragraph{Omitted variables:}
Some variables defined in the URScript are not used in the model to minimize the number of variables. For instance, current and target positions are not included if they are only used to calculate a distance. Instead, the distance is input directly. Adopting a similar approach, other variables are omitted or given a boolean rather than a numerical type to reduce the state space. 

\subsection{RoboChart Model}
\label{section:robochart}

In writing a RoboChart model, a key decision is the definition of events and operations that capture services of the robotic platform. The previous section describes our assumptions, some of which are reflected in these definitions. These services are not further specified and establish the interface of the model.  Properties are described in terms of interaction with the modelled component via these services.  

\begin{figure}[!hbt]
    \centering
    \includegraphics[width=\textwidth]{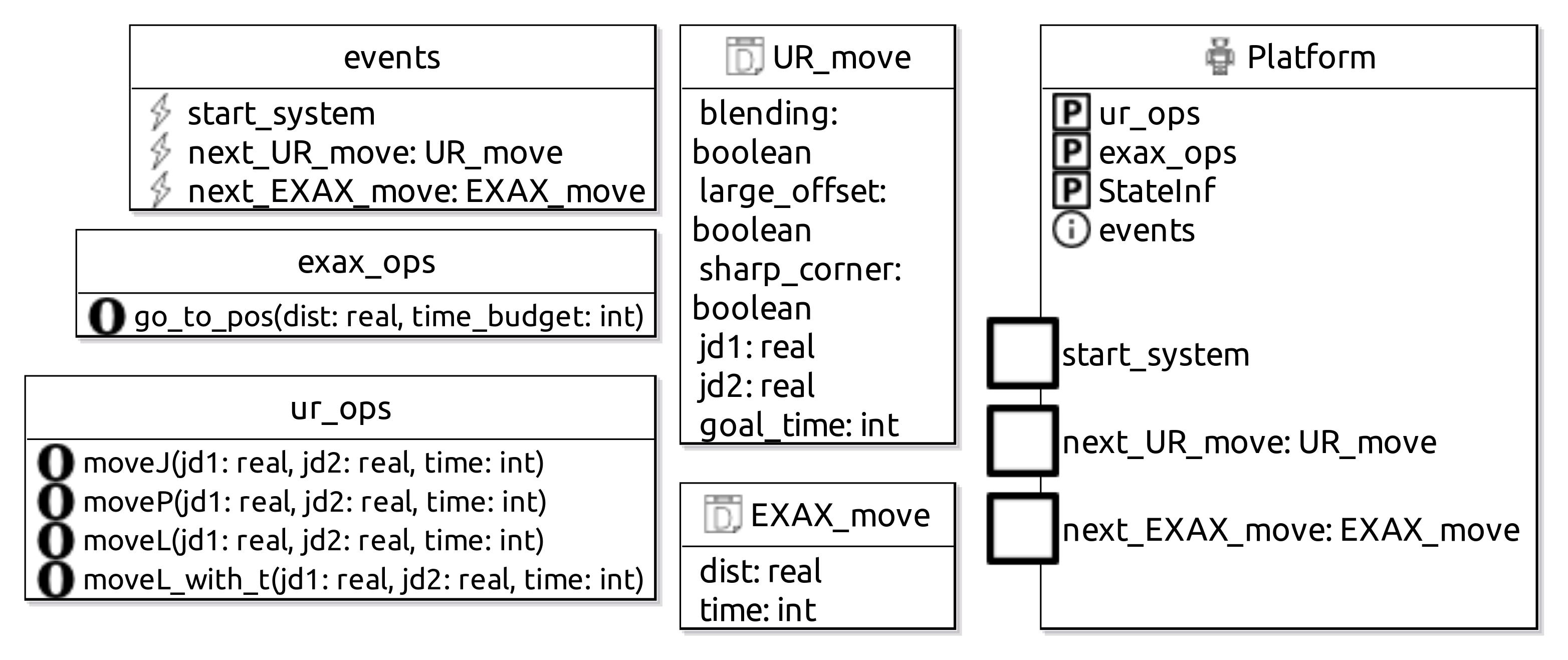}
    \caption{Robotic platform with its defined events, provided operations and custom record types to represent move commands.}
    \label{fig:platform}
\end{figure}

Fig.~\ref{fig:platform} shows the robotic platform of our model~(on the right), with three input events~(\RC{start_system}, \RC{next_UR_move} and \RC{next_EXAX_move})  and the five operations that can be called~(four operations in \RC{ur_ops} and one in \RC{exax_ops}). The events, declared in the \RC{events} interface, are used to initiate the system and send new move requests as previous moves are completed. The operations are defined in two separate interfaces:~\RC{ur\_ops} and \RC{exax\_ops}, corresponding to the move commands that are executed by the UR robot and the Carpano FIVE.  Although the model declares one robotic platform, it captures services of both the robot and the turntable used by the software.

\begin{figure}[!hbt]\centering
    \includegraphics[width=\textwidth, keepaspectratio]{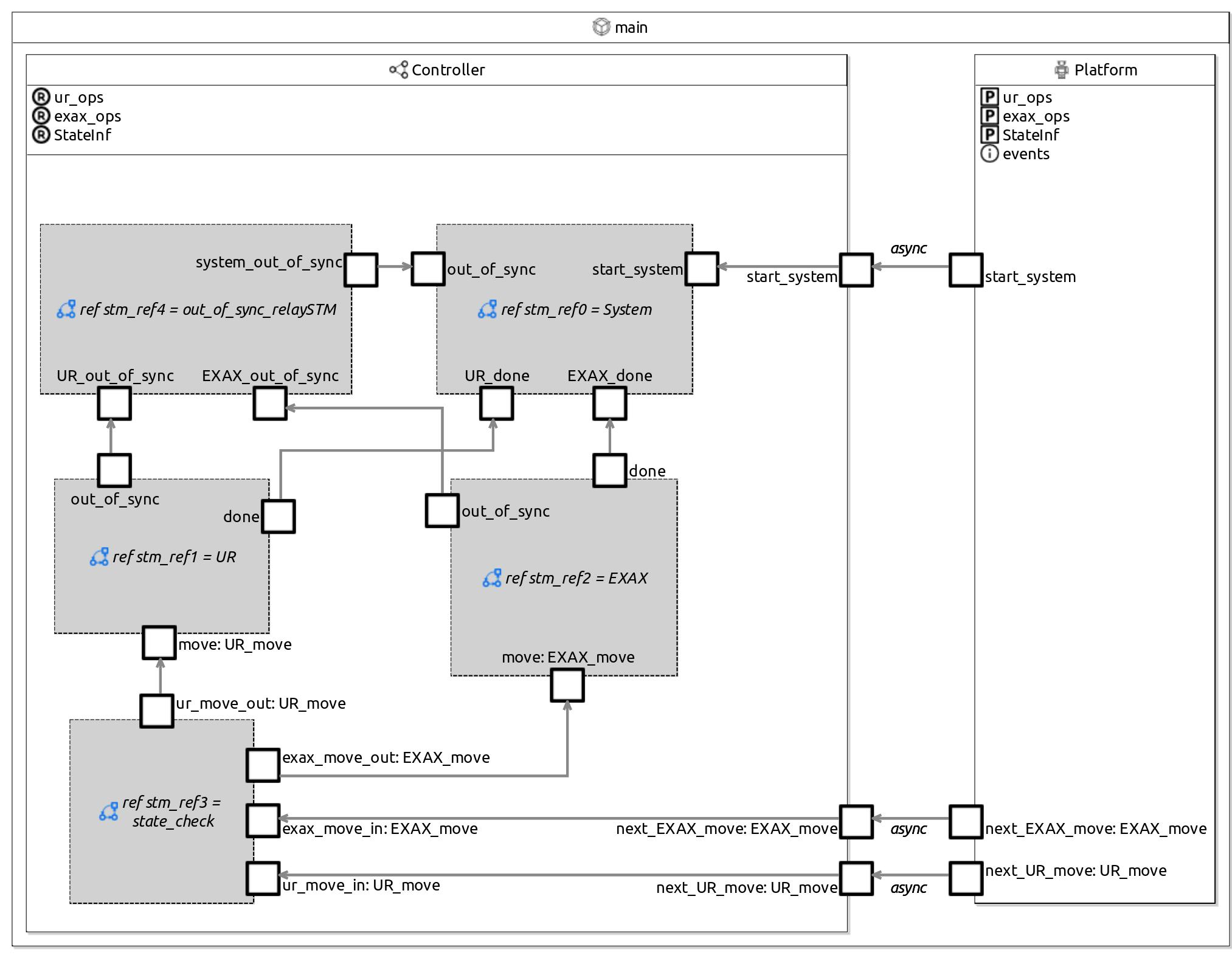}
    \caption{Main module including the Controller with all state machines and the robotic platform.}
    \label{fig:module}
\end{figure}

\paragraph{Module}
A RoboChart model is defined by a module block, including the robotic platform and, in our case, one controller block, as shown in Fig.~\ref{fig:module}. Our module is called \RC{main}, and the behaviour of our controller is defined by five parallel state machines, acting over the events and operations of the robotic platform as declared in three required interfaces and reflected in connections between the robotic platform and the controller~(arrows between the blocks annotated with \RC{async}). 

A RoboChart controller defines how the events of the robotic platform connect to its state machines.  In our example, \RC{start\_system} is used by the machine called \RC{System}. The other two events are used by the machine \RC{state\_check}.  A controller also defines how its state machines are connected to each other, via their events, to exchange information and synchronise their behaviour. In Fig.~\ref{fig:module}, the \RC{Controller} block includes five blocks, each a reference to one of its state machines, as indicated by the keyword \RC{ref}. In what follows, we present the definition of these machines.  

\begin{figure}[!hbt]
    \centering
    \includegraphics[scale=0.17]{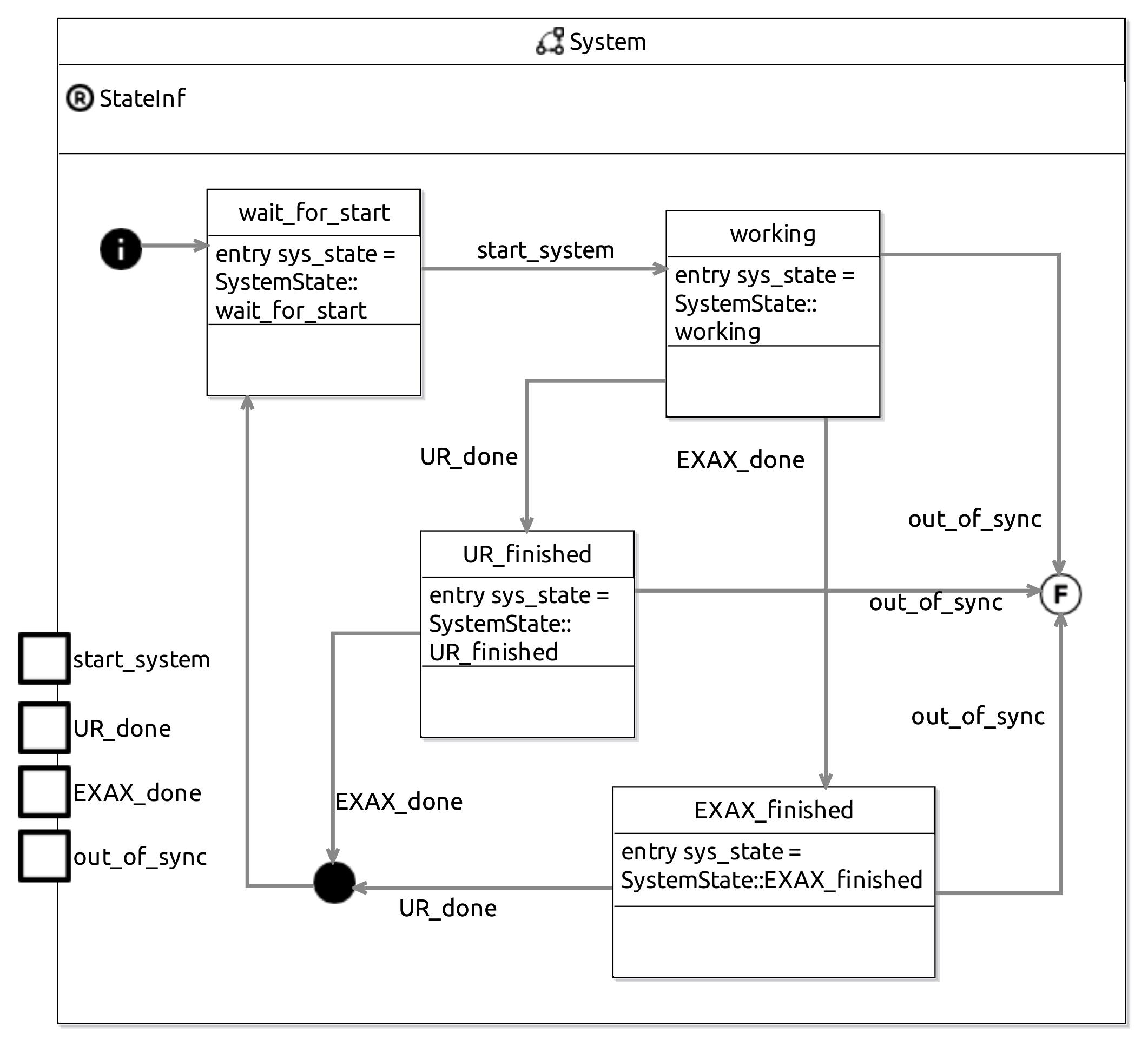}
    \caption{\RC{System} state machine.} 
    \label{fig:systemSTM}
\end{figure}

\paragraph{The \RC{System} state machine} Its definition is shown in Fig.~\ref{fig:systemSTM}. In each state of \RC{System}, a shared variable \RC{sys\_state} is updated to record the current state of the system.  This variable is used in the \RC{state_check} machine presented later to decide whether or not a movement request should be forwarded to the \RC{EXAX} or to the \RC{UR} state machine, that is to the turntable or to the UR robot. 

The initial junction, a black circle with an \RC{i}, indicates \RC{wait\_for\_start} as the initial state of \RC{System}, where it  waits for the event \RC{start\_system}. When \RC{start\_system} happens, \RC{System} moves to the \RC{working} state, where it stays until either the UR robot or the EXAX finishes all of their waypoints, as indicated by events \RC{UR\_done} and \RC{EXAX\_done}, or an \RC{out\_of\_sync} event occurs. An \RC{out\_of\_sync} event, from any of the states \RC{working}, \RC{UR\_finished} or \RC{EXAX\_finished}, results in a transition to the \RC{final} state: white circle with an \RC{F}. This means that the \RC{System} state machine cannot progress further. 

If both \RC{UR\_done} and \RC{EXAX\_done} occur, regardless of in which order, \RC{System} goes through either the state \RC{UR_finished} (if the UR robot finishes first) or \RC{EXAX_finished} (if the EXAX finishes first), before going back to \RC{wait\_for\_start}. This is the end of a welding operation. 

\begin{figure}[!hbt]
    \centering
\includegraphics[width=\textwidth, keepaspectratio]{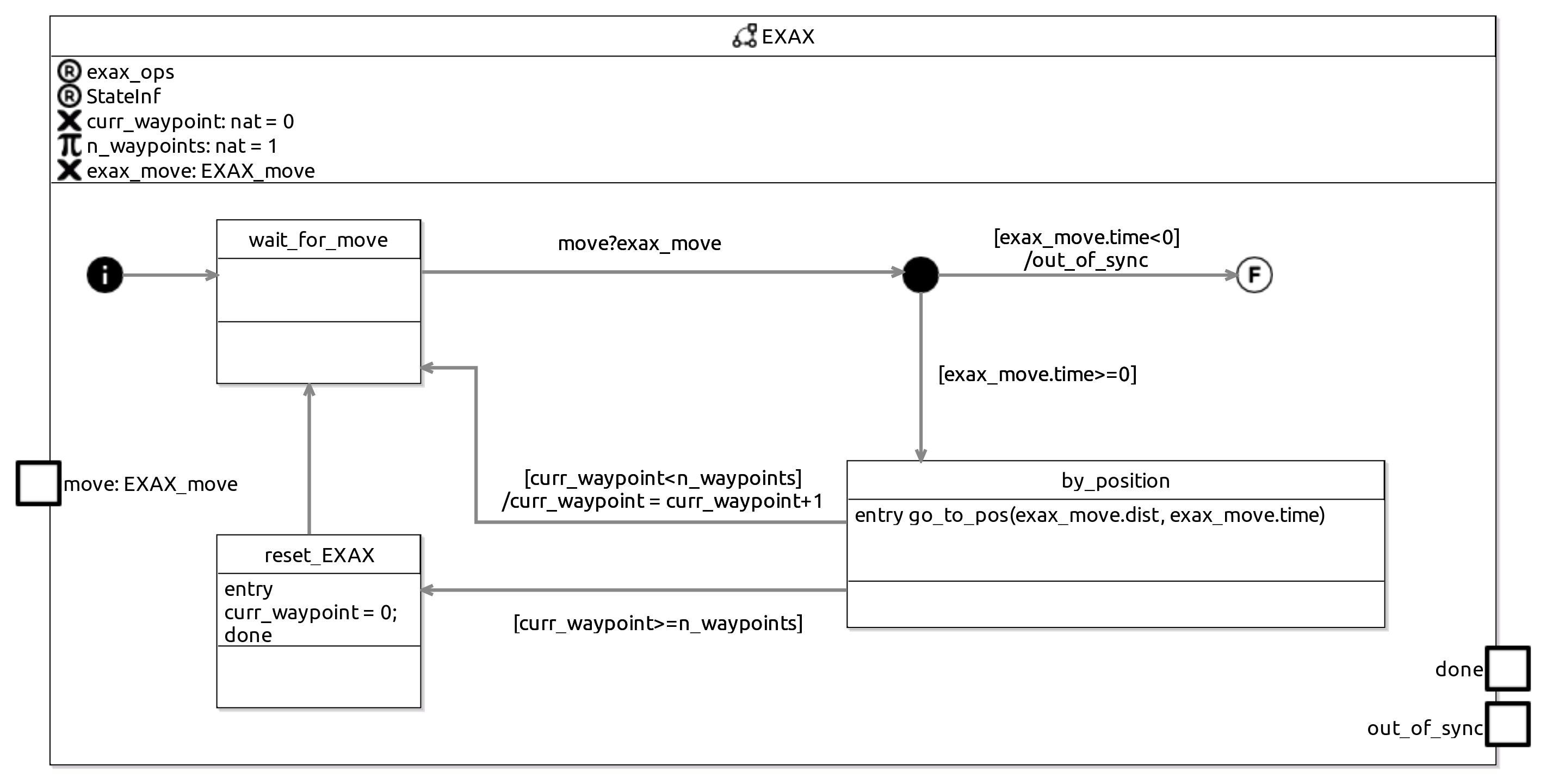}
    \caption{EXAX state machine}
    \label{fig:EXAX_stm}
\end{figure}

\paragraph{The \RC{EXAX} state machine} Its definition, shown  in 
Fig.~\ref{fig:EXAX_stm}, captures the behaviour of the turntable. \RC{EXAX} starts in the \RC{wait\_for\_move} state, waiting for a \RC{move} command. The variable \RC{curr\_waypoint} is initialised to 0, and with the constant \RC{n\_waypoints} defined as 1, as in Fig.~\ref{fig:EXAX_stm}, the turntable goes through two waypoints. When \RC{EXAX} receives a \RC{move} event, it stores the requested distance and time to move in a variable  \RC{exax\_move}. In the junction~(dark circle), it is checked if the time of the movement request~(\RC{exax\_move.time}) is strictly negative. If it is, an \RC{out\_of\_sync} event is triggered and \RC{EXAX} terminates. Otherwise, \RC{EXAX} moves to a state \RC{by\_position}, where the operation \RC{go\_to\_pos} is called using as arguments the values in \RC{exax\_move}. If \RC{curr\_waypoint} is greater or equal to \RC{n\_waypoints}, \RC{curr\_waypoint} is reset and the \RC{done} event is triggered. This is then relayed, by the controller, to the \RC{System} state machine via its event \RC{EXAX\_done}. (\RC{System} then transitions to its state \RC{EXAX\_finished}).

\begin{figure}[!hbt]
    \centering
    \includegraphics[width=\textwidth, keepaspectratio]{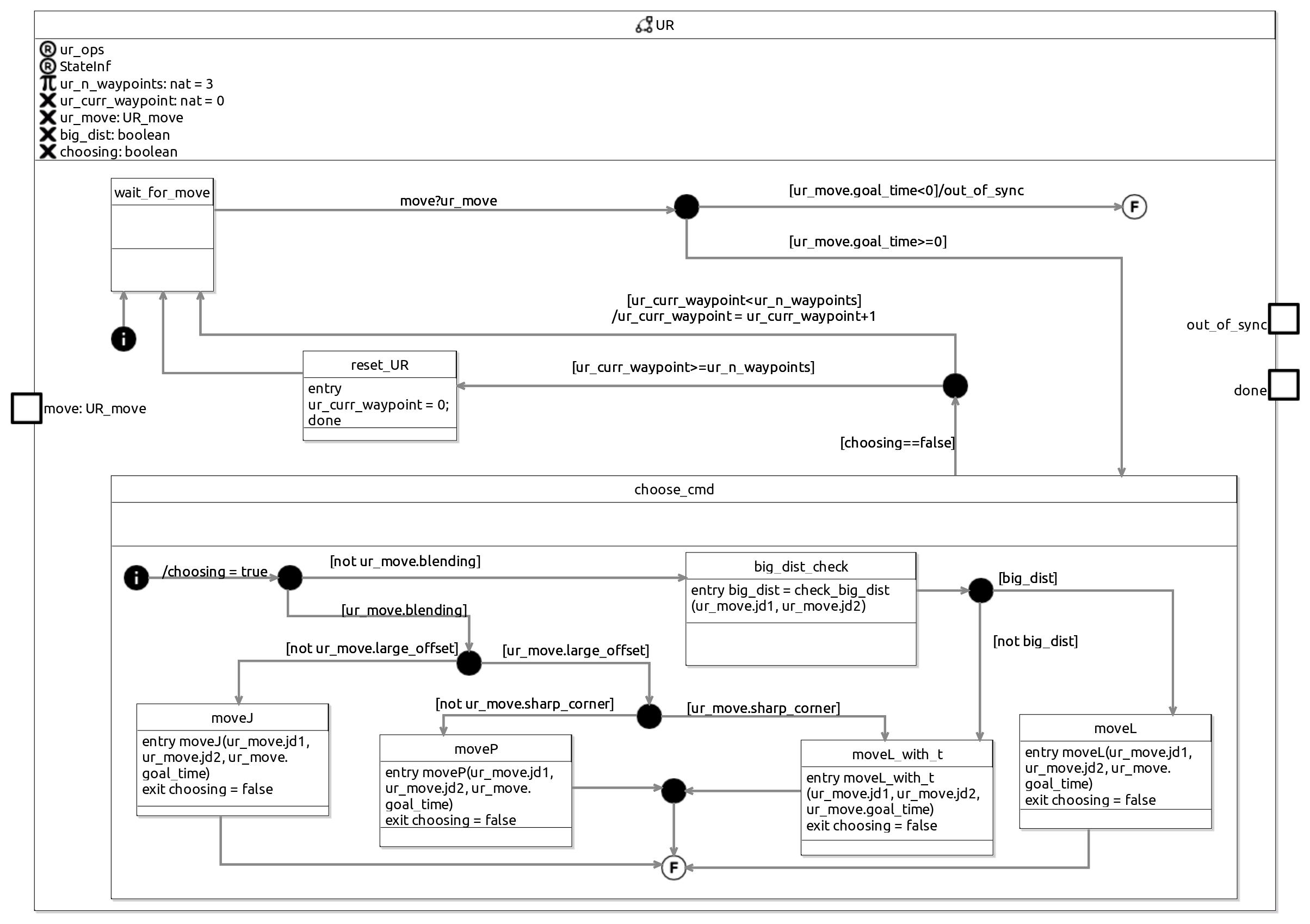}
    \caption{UR state machine.}
    \label{fig:UR_stm}
\end{figure}

\paragraph{The \RC{UR} state machine} It is defined in 
Fig.~\ref{fig:UR_stm} to model the behaviour of the UR robot, and is similar to  \RC{EXAX}~(Fig.~\ref{fig:EXAX_stm}).
The same method of counting and incrementing waypoints is used, and the \RC{done} and \RC{out\_of\_sync} events are used in the same way. The UR robot, however, chooses the most suitable movement type, so \RC{choose\_cmd} is more complex than the \RC{by\_position} state of \RC{EXAX}. 

Upon entering the \RC{choose\_cmd} state, the boolean variable \RC{choosing} is set to true. This ensures that a \RC{move} command must be chosen before leaving the state, since the only transition out of the state \RC{choose\_cmd} has a guard that requires the value of \RC{choosing} to be \RC{false}. 

Which move command is chosen depends on whether or not the move request includes blending, a large offset from the ideal path (set to 0.8mm in our use case), or a sharp corner. The first junction checks whether the move request includes a blend radius or not, that is, whether \RC{ur\_move.blending} is \RC{true} or \RC{false}, where the variable \RC{ur\_move} records the data associated with the \RC{move} request as defined in the transition out of \RC{wait\_for\_move}. If it does include a blend radius, the next junction checks whether the move request includes a large offset~(\RC{ur\_move.large\_offset}). 

If the offset is smaller than the threshold, \RC{moveJ} is chosen:~\RC{UR} transitions to the \RC{moveJ} state, where the operation of the same name is called and the variable \RC{choosing} is set to \RC{false} in the \RC{exit} action.  With that \RC{choose\_cmd} is exited. If the offset is large, the next junction checks whether the move request contains a sharp corner or not~(\RC{ur\_move.sharp\_corner}). If it does not, \RC{moveP} is suitable, but if it does, it is necessary to use \RC{moveL_with_t}. In each case, like for \RC{moveJ}, the \RC{entry} action of a state calls the right operation and the \RC{exit} action updates \RC{choosing}. If the move command does not include blending, the system enters the \RC{big_dist_check} state after the very first junction.  In the \RC{entry} action of that state, the \RC{check\_big\_dist} function checks if the absolute value of either of the joint distances is larger than a given value~(in the example, 1), since the distance determines if \RC{moveL} is sufficient. If the distance is short, a \RC{moveL\_with\_t} command is issued in the state of the same name. 

\paragraph{The \RC{out\_of\_sync Relay} state machine} It is simple and omitted here; it relays the \RC{out\_of\_sync} event from the \RC{EXAX} and \RC{UR} to \RC{System}. This is necessary just because RoboChart prohibits connecting two different events to the same input of another machine. Full details can be found in~\cite{NordlieThesis}. 

\begin{figure}[htbp!]
    \centering
    \includegraphics[width=\textwidth, keepaspectratio]{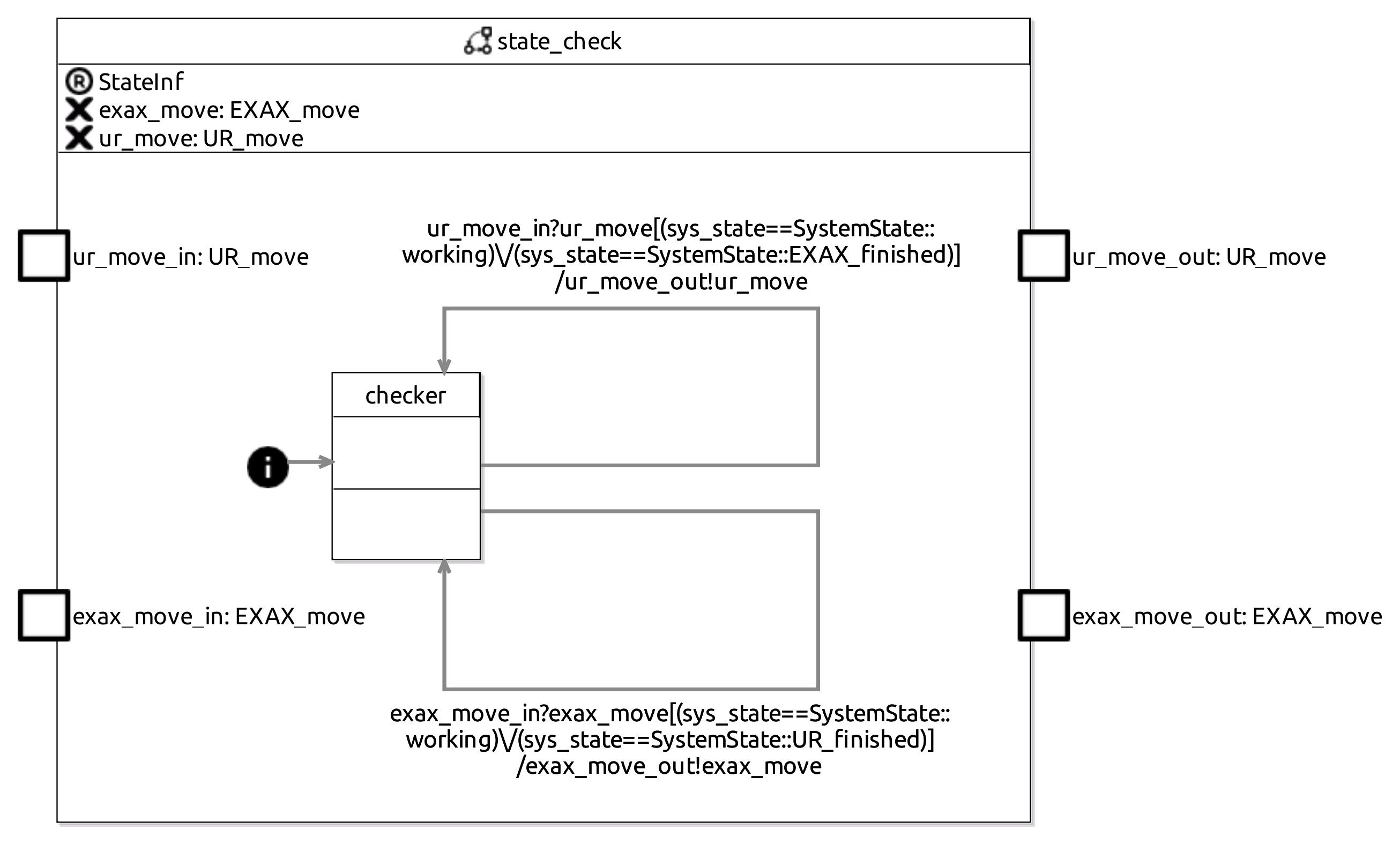}
    \caption{State machine responsible for relaying the move commands of the \RC{UR} and \RC{EXAX} only if \RC{System} is in a state where those state machines should receive commands.}
    \label{fig:state_check}
\end{figure}

\paragraph{The \RC{state\_check} state machine}
It is defined in Fig.~\ref{fig:state_check} and has a single state \RC{checker} with two self-transitions. They are triggered by events that accept and record a \RC{move} command in local variables \RC{ur\_move} or \RC{exax\_move} depending on whether the \RC{UR} or the \RC{EXAX} received a move request~(whether an input event \RC{ur\_move\_in} or \RC{exax\_move\_in} happens).  

The guards of the transitions ensure that these inputs are accepted only if the system is in a state where the move command should be forwarded to the \RC{UR} or \RC{EXAX} machines. There are only two states where they should move:~states \RC{working} or \RC{EXAX\_finished}, for the event \RC{ur\_move\_in}, and \RC{working} or \RC{UR\_finished}, for \RC{exax\_move\_in}. In the actions of the transitions, if a move request for the UR robot arrives, it is forwarded to the \RC{UR} state machine. Similarly, a move request for EXAX is forwarded to the \RC{EXAX} state machine. With the guards in the transitions, the \RC{state\_check} machine ensures that no move operations can be executed before the system has started, and that once a robot has reached all its waypoints, no further move operations can be executed until the system is reset and restarted.

\section{Model Checking}
\label{sec:modelchecking}

In this section, we describe the model checking and its results:~the defined properties and assertions~(Section~\ref{sec:properties}), the results from FDR~(Section~\ref{sec:results}), and their implications for the real-life system~(Section~\ref{sec:reallife}).

\subsection{Verification of Selected Properties}
\label{sec:properties}
Based on the requirements from Section~\ref{sec:specs}, assertions can be formulated in natural language, and later defined in $tock$-CSP, which is a dialect of the process algebra CSP where the event $tock$ marks the passage of discrete time.
To this end, the following properties are to be validated through model checking~\cite{NordlieThesis}:
\begin{itemize}
    \item Every time an \RC{EXAX\_move}  or \RC{UR\_move} input event is triggered by the robotic platform, the corresponding movement operation for \RC{EXAX} or \RC{UR}, respectively, is called. This is captured in \lstinline{assertion A1} and \lstinline{A2} for \RC{EXAX} , and in \lstinline{assertion A3} and \lstinline{A4} for \RC{UR}.
    \item The \RC{EXAX} state machine and the \RC{UR} state machine do not terminate. This is captured in \lstinline{assertion A5}
    and in \lstinline{assertion A6}, respectively.
    \item If no \RC{out\_of\_sync} event occurs in the \RC{System} state machine, the state machine does not terminate. This is captured in \lstinline{assertion A7}.
\end{itemize}
All the assertions described above are detailed next. 

\paragraph{Assertion A1}
We present in Listing~\ref{csp:SpecA1} the RoboTool script defining \lstinline{assertion A1}. With that, RoboTool can use FDR to check whether the assertion holds or not. FDR uses $tock$-CSP processes that define the semantics of the state machines presented in the previous section.  These $tock$-CSP processes are automatically calculated by RoboTool. The scripts are written in a mixture of natural language and CSP.

As defined in Listing~\ref{csp:SpecA1}, \lstinline{assertion A1} requires that \lstinline {EXAX refines SpecA1 /@in the traces model@/}, on line 8. This means that \lstinline{assertion A1} requires the traces of the process \lstinline{EXAX} for the machine of the same name to be also traces of the process \lstinline{SpecA1} defined in lines~1-7. 

\begin{lstfloat}[htbp!]
\begin{lstlisting}
timed csp  SpecA1 /@csp-begin@/
Timed (OneStep) {
	SpecA1 = let
		      Def = (CHAOS(Events) [| {|EXAX::move.in|} |> 
        ADeadline({|EXAX::go_to_posCall|}, 0)); Def
	within timed_priority(Def) }
/@csp-end@/
timed assertion A1: EXAX refines SpecA1 /@in the traces model@/.
\end{lstlisting}
\caption{\label{csp:SpecA1} Definition of \lstinline{SpecA1} and \lstinline{assertion A1}.}
\end{lstfloat}

\lstinline{SpecA1} is defined directly in CSP, as indicated in lines~1 and 7. 
 Moreover, it is defined within a \lstinline{Timed} section~(line~2). So, it is a $tock$-CSP process. It is given by the equation in line~3.
 
The definition of \lstinline{SpecA1} uses a \lstinline{let}-\lstinline{within} construct. In the \lstinline{let} clause, a process \lstinline{Def}~(lines~4-5) is defined. In the \lstinline{within} clause, it is used to define \lstinline{SpecA1} using a \lstinline{timed_priority} function.  This is just a technicality of FDR:~\lstinline{timed_priority} enforces the understanding of $tock$ as a special event that marks the passage of time. So, the behaviour of \lstinline{SpecA1} is really that of \lstinline{Def}.

The behaviour of \lstinline{Def} is initially that of \lstinline{CHAOS(Events)}, a process that allows any event to occur. It can, however, be interrupted~(operator \lstinline{[| ... |>]}) by the CSP event \lstinline{EXAX::move.in}, which represents the input \RC{move}.  Upon interruption, the behaviour of \lstinline{Def} is give by \lstinline{ADeadline}. This is a parameterised CSP process defined in the RoboTool $tock$-CSP mechanisation~\cite{BRC22}  that takes a set of events and a deadline, given as a number of $tock$ events, as arguments. It requires that one of the events in the provided set, in this case only \lstinline{EXAX::go_to_posCall}, the CSP process representing a a call to \RC{go\_to\_pos}, occurs within the deadline, which here is set to 0. Thus, the call to the \RC{go\_to\_pos} operation is required to happen immediately when the \lstinline{EXAX::move_in} event occurs.

\paragraph{Assertion A2}
For \lstinline{assertion A1} to be meaningful, it is necessary to ensure that the \RC{EXAX} state machine is timelock-free, due to the trivial case where the process refuses the event $tock$. This is checked with  \lstinline{assertion A2}.  Listing~\ref{csp:A2} shows the definition of \lstinline{A2}, where \lstinline{EXAX2} on line 3 is defined as a version of \lstinline{EXAX} in whose traces the events \lstinline{EXAX::go_to_posCall} are ignored (using the hidden operator:~\lstinline{\ }). This is done because the machine can timelock in the call to that operation, that is, refuse the $tock$ event. This is because that call, being in an \RC{entry} action, is urgent, and deadlines create potential timelocks. In $tock$-CSP, when a deadline is reached, $tock$ is refused. For the \lstinline{EXAX::D__} process~(line 3), two arguments (0, 1) are needed due to technicalities of the CSP model of RoboChart. The first argument is an ID-value, and the second is the value of \RC{n\_waypoints}, which is not fixed in the model of a machine.  

  \begin{lstfloat}[htbp!]
\begin{lstlisting}
timed csp  EXAX2 /@csp-begin@/
Timed(OneStep) {
  EXAX2 = EXAX::D__(0, 1) \ {| EXAX::go_to_posCall |}
}
/@csp-end@/
assertion A2: EXAX2 is timelock-free.
\end{lstlisting}
\caption{\label{csp:A2} Definition of \lstinline{EXAX2} and \lstinline{assertion A2}.}
\end{lstfloat}

\paragraph{Assertion A3}
The definition of \lstinline{assertion A3} and \lstinline{SpecA3} can be seen in Listing~\ref{csp:SpecA3+A3}. This is the UR equivalent to \lstinline{assertion A1} and \lstinline{SpecA1}. This assertion ensures that for each \RC{move} event, one of the four move operation calls must be made before any time is allowed to pass.

\begin{lstfloat}[htbp!]
\begin{lstlisting}
timed csp  SpecA3 /@csp-begin@/
Timed(OneStep) {
  SpecA3 = let
    Def = (CHAOS(Events) [| {|UR::move.in|} |> ADeadline(
      {|UR::moveJCall,UR::movePCall,UR::moveLCall, UR::moveL_with_tCall|},0)); 
    Def
within timed_priority(Def) }
/@csp-end@/
timed assertion A3: UR refines SpecA3 in the traces model.
\end{lstlisting}
\caption{\label{csp:SpecA3+A3} Definition of \lstinline{SpecA3} and \lstinline{assertion A3}}
\end{lstfloat}

\paragraph{Assertion A4}
Also for the \RC{UR} STM it is important to ensure timelock-freedom, and this is done in \lstinline{assertion A4}, which is the UR equivalent to \lstinline{assertion A2} and omitted here. 


\paragraph{Assertions A5 and A6}
%
They are defined in Listing~\ref{csp:A5}; they require that the \RC{EXAX} and \RC{UR} state machines, respectively, do not terminate. These assertions are expected to pass given that no \RC{out\_of\_sync} occurs.
\begin{lstfloat}[htbp!]
\begin{lstlisting}
assertion A5: EXAX /@does not terminate@/.
assertion A6: UR /@does not terminate@/.
\end{lstlisting}
\caption{\label{csp:A5} Definition of \lstinline{assertion A5} and \lstinline{assertion A6}.}
\end{lstfloat}

\paragraph{Assertion A7}
Listing~\ref{csp:A7} shows the definition of \lstinline{SysTerminates}, as well as the definition of a process \lstinline{Stop}, and \lstinline{assertion A7}. The process \lstinline{SysTerminates}~(line~3) is based on another process (\lstinline{SysConstrained} from line 2) which is a version of the system where \RC{out\_of\_sync} events are skipped. So, the \lstinline{SysTerminates} process on line~3 only takes into account the termination event of the \RC{System} state machine. This is done by hiding all events except \lstinline{System::terminate} using the \lstinline{|\} operator. If \RC{System} terminates despite the \RC{out\_of\_sync} event being ignored, the assertion should fail, and this is captured by comparing \lstinline{SysTerminates} to the process \lstinline{Stop}~(line 6-8). \lstinline{Stop} is equivalent to the CSP process \lstinline{STOP}, which is a deadlock. This means that the \lstinline{Stop} process can never perform any events before terminating, and so by demanding that \lstinline{SysTerminates refines Stop /@in the traces model@/}, it can be ensured that this process never performs \lstinline{System::terminate}, and thus never terminates. The assertion is expected to always pass since the \RC{out\_of\_sync} event is being ignored. Still, it shows that in the cases where it does not occur, the \RC{System} state machine does not terminate.  

\begin{lstfloat}[htbp!]
\begin{lstlisting}
timed csp  Systerminates associated to System /@csp-begin@/
	SysConstrained = (System::D__(0) [| {| System::out_of_sync |} |] SKIP)
	SysTerminates = (SysConstrained ; System::terminate -> SKIP) |\ {| System::terminate |}
/@csp-end@/

csp Stop /@csp-begin@/
    Stop = STOP
/@csp-end@/

assertion A7: SysTerminates refines Stop in the traces model.
\end{lstlisting}
\caption{\label{csp:A7} Definition of \lstinline{SysTerminates}, \lstinline{Stop} and \lstinline{assertion A7}.}
\end{lstfloat}

\subsection{Results from Checking the Assertions}
\label{sec:results}

As previously mentioned, the assertions have been run with two different value ranges for the time variable, \RC{core\_int}. The range \RC{[0..2]} implies that it is not possible to receive negative time budgets for the movements, meaning that the movements are always performed in accordance with nominal plans. The range \RC{[-1..1]} implies that negative time budgets can occur, signifying either an infeasible plan from Delfoi or incorrect execution of movements by either the UR robot or the turntable. The assertions have been checked on a computer with an AMD Dual EPYC 7501 (2*32 cores) processor and 2TiB of RAM.

\paragraph{Nominal case with only positive time budgets, \RC{[0..2]}}
\label{section:nominal}
As summarized in Table~\ref{tab:positive-time}, all assertions pass. 
This outcome is desirable for verifying the synchronisation properties. Since \lstinline{A1} and \lstinline{A3} pass, it can be concluded that a movement operation is always called for each movement request received. Relating back to the requirements in Section~\ref{sec:specs}, it indicates that \textbf{R2} is satisfied.

\begin{table}
\centering
\begin{tabular}{|l|c|l|l|l|l|l|}
\hline
\multicolumn{1}{|c|}{\multirow{2}{*}{\textbf{Assertion}}} 
& \multicolumn{1}{c|}{\multirow{2}{*}{\textbf{Result}}} 
& \multicolumn{3}{c|}{\textbf{Elapsed Time}} 
& \multicolumn{2}{c|}{\textbf{Complexity}} 
\\ \cline{3-7} 
\multicolumn{1}{|c|}{}
& \multicolumn{1}{c|}{}
& \multicolumn{1}{c|}{\textbf{Compilation}}
& \multicolumn{1}{c|}{\textbf{Verification}}
& \multicolumn{1}{c|}{\textbf{Total}}
& \multicolumn{1}{c|}{\textbf{States}} 
& \multicolumn{1}{c|}{\textbf{Transitions}} 
\\ \hline
{\lstinline!A1!}  & $\checkmark$  & 10.79s & 0.34s & 11.13s     & 228 & 713         \\ \hline
{\lstinline!A2!}  & $\checkmark$  & 11.10s & 0.32s  & 11.42s     & 228  & 713           \\ \hline
{\lstinline!A3!}  & $\checkmark$  & 14.42s & 0.51s  & 14.93s     & 5,060   & 17,001           \\ \hline
{\lstinline!A4!}  & $\checkmark$  & 14.15s & 0.57s  & 15.72s     & 5,060   & 17,001           \\ \hline
{\lstinline!A5!}  & $\checkmark$  & 0.12s  & 0.39s & 0.51s   & 228  &  713             \\ \hline
{\lstinline!A6!}  & $\checkmark$  & 0.12s  & 0.60s & 0.72s   & 5,060   & 17,001                 \\ \hline
{\lstinline!A7!}  & $\checkmark$  & 0.64s  & 0.55s & 1.19s   & 32   & 197          \\ \hline
\end{tabular}
\cprotect\caption{Results of model-checking all assertions in FDR with \RC{core\_int = [0..2]}.}
\label{tab:positive-time}
\end{table}

\begin{figure}[htbp!]
    \centering
    \includegraphics[width=\textwidth, keepaspectratio]{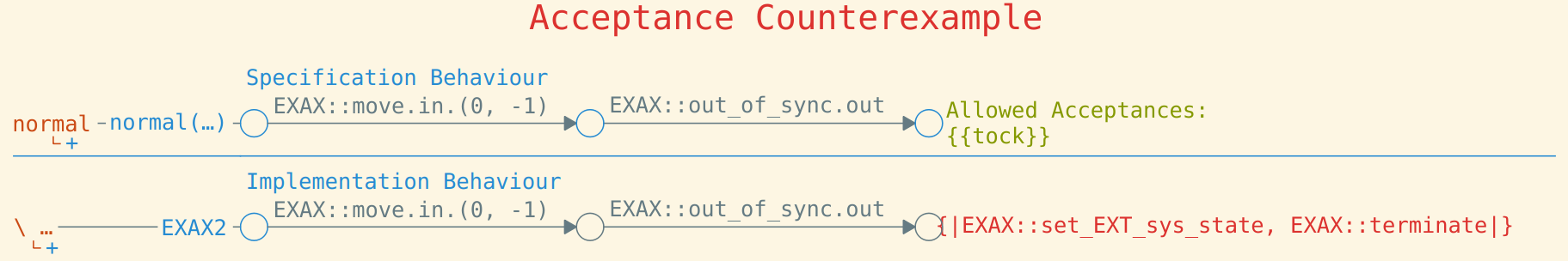}
    \caption{Trace showing a counterexample to assertion \lstinline{A5} - \lstinline{EXAX does not terminate}.}
    \label{fig:EXAX_terminates_trace}
\end{figure}

\paragraph{Realistic case with possibility of negative time budget, \RC{[-1..1]}}
\label{section:failure}
As summarised in Table~\ref{tab:negative-time}, the only assertion that passes is \lstinline{A7}. 
\begin{table}
\centering
\begin{tabular}{|l|c|l|l|l|l|l|}
\hline
\multicolumn{1}{|c|}{\multirow{2}{*}{\textbf{Assertion}}} 
& \multicolumn{1}{c|}{\multirow{2}{*}{\textbf{Result}}} 
& \multicolumn{3}{c|}{\textbf{Elapsed Time}} 
& \multicolumn{2}{c|}{\textbf{Complexity}} 
\\ \cline{3-7} 
\multicolumn{1}{|c|}{}
& \multicolumn{1}{c|}{}
& \multicolumn{1}{c|}{\textbf{Compilation}}
& \multicolumn{1}{c|}{\textbf{Verification}}
& \multicolumn{1}{c|}{\textbf{Total}}
& \multicolumn{1}{c|}{\textbf{States}} 
& \multicolumn{1}{c|}{\textbf{Transitions}} 
\\ \hline
{\lstinline!A1!}  & $\RC{X}$  & 11.16s & 0.26s & 11.42s     & 11 & 87              \\ \hline
{\lstinline!A2!}  & $\RC{X}$  & 10.94s & 0.28s  & 11.22s     & 99  & 388           \\ \hline
{\lstinline!A3!}  & $\RC{X}$  & 15.23s & 0.26s  & 15.49s     & 75   & 1,235           \\ \hline
{\lstinline!A4!}  & $\RC{X}$  & 14.19s & 0.35s  & 14.54s     & 931   & 4,412           \\ \hline
{\lstinline!A5!}  & $\RC{X}$  & 0.10s  & 0.40s & 0.50s   & 153  &  554             \\ \hline
{\lstinline!A6!}  & $\RC{X}$  & 0.12s  & 0.48s & 0.60s   & 1,497   & 6,329                 \\ \hline
{\lstinline!A7!}  & $\checkmark$  & 0.70s  & 0.58s & 1.28s   & 32   & 197           \\ \hline
\end{tabular}
\cprotect\caption{Results of model-checking all assertions in FDR with \RC{core\_int = [-1..1]}.}
\label{tab:negative-time}
\end{table}
Fig.~\ref{fig:EXAX_terminates_trace} shows an example of a trace related to the failed assertion \lstinline{A5}. It shows an \RC{out\_of\_sync} event, which should lead to termination due to \RC{EXAX\_move} having a negative value for the time variable.
The counterexample, given at the bottom in Fig.~\ref{fig:EXAX_terminates_trace}, shows that the system does not terminate. 
Since \RC{out\_of\_sync} events are possible, it shows that \textbf{R1} from Section~\ref{sec:specs} is satisfied.

\subsection{Implications and Real-Life System Improvements}
\label{sec:reallife}

\begin{figure}[!ht]
   \centering
 \includegraphics[scale=0.3]{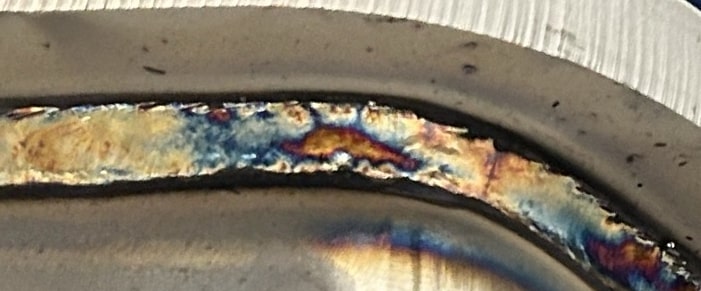}
 \caption{A corner of the workpiece showing significantly improved welding quality after system re-calibration.}
 \label{fig:WeldRecalib}
\end{figure}

The fact that all assertions in the nominal case (\RC{core\_int = [0..2]}) pass, and all assertions apart from one in the realistic case  (\RC{core\_int = [-1..1]}) fail, indicates that the robots are unable to follow the nominal plans. This could be, for instance, due to hardware limitations, like insufficient maximum speed or acceleration, or inaccuracy in their trajectory following. To minimise errors, a full re-calibration of the real-life system has been done. Since the programming is offline in Delfoi, it is crucial that the physical system is calibrated precisely so the digital CAD model is correctly positioned and orientated. The resulting weld after the calibration, as seen in Fig.~\ref{fig:WeldRecalib}, shows a significant improvement in quality. 

\section{Conclusion and Further Work}
\label{sec:conclusion}

Applying model checking to an already existing industrial robotic system with known weaknesses has proved to be both challenging and useful. The main challenge lies in ensuring that the model catches the essential characteristics of the real-life system. Abstractions and assumptions need to be made to keep the computational complexity at a reasonable level. However, it is crucial that they are not so limiting that the model fails to capture the behaviour and possible errors. Keeping an eye on this so-called ``reality gap'' between the model and the real system is vital.

Even though improvements have been made on the real-life system based on the findings from the model checking, there is still room for improvement. 
An interesting path for further work is to verify the assumptions made on the hardware, to achieve a co-verification similar to that in~\cite{murray_safety_2022}. It is also beneficial to verify the offline programming in Delfoi, and set requirements for the generation of waypoints. This will ensure the feasibility of the planned trajectories.


Further work will use the model further along the RoboStar workflow in~\cite{cavalcanti_robostar_2021}. The next step is to automatically generate a simulation model RoboSim~\cite{CSMRCDLT19}. The RoboStar team is also working on model-based testing, so we can generate forbidden traces from RoboChart models. They are specifications of tests that can then be run against the real-life software. This is an interesting way to address the reality gap.

\section*{Acknowledgements}  
David Anisi has received partial funding from the Norwegian Research Council (RCN) RoboFarmer, project number 336712. Ana Cavalcanti and Pedro Ribeiro are funded by the Royal Academy of Engineering~(Grant No CiET1718/45), and the UKRI~(UK Research and Innovation Council), Grants No~EP/R025479/1 and EP/V026801/1.

\newpage
\nocite{*}
\bibliographystyle{eptcs}
\bibliography{IntelliWelder}
\end{document}